\documentclass[runningheads]{llncs}

\usepackage{eccv}

\usepackage{eccvabbrv}

\usepackage{graphicx}
\usepackage{booktabs}

\usepackage[accsupp]{axessibility}  

\usepackage{hyperref}

\usepackage{orcidlink}

\usepackage{amsmath}
\usepackage{multirow}
\usepackage{float}
\usepackage{hyperref}

\begin{document}

\title{Enhancing Dataset Distillation via Label Inconsistency Elimination and Learning Pattern Refinement} 

\titlerunning{Track 1 No.1 Solution: M-DATM}

\author{
Chuhao Zhou\inst{1}\orcidlink{0000-0002-4363-8931}\index{Zhou, Chuhao} \and
Chenxi Jiang\inst{1}\orcidlink{0009-0005-5581-097X}\index{Jiang, Chenxi} \and Yi Xie\inst{3}\orcidlink{0000-0002-1938-4089}\index{Xie, Yi} 
\and Haozhi Cao\inst{2}\orcidlink{0000-0001-7703-3490}\index{Cao, Haozhi}
\and \\
Jianfei Yang\inst{1,2}\thanks{J. Yang is the project lead.}\orcidlink{0000-0002-8075-0439}\index{Yang, Jianfei}
}

\authorrunning{C.~Zhou et al.}

\institute{
School of Mechanical and Aerospace Engineering, NTU, Singapore \\
\and
School of Electrical and Electronic Engineering, NTU, Singapore \\
\and
Duke-NUS Medical School, NUS, Singapore\\
\email{\{chuhao002, chenxi003, haozhi002\}@e.ntu.edu.sg \\ yi.xie@u.duke.nus.edu,jianfei.yang@ntu.edu.sg \\
      }
}

\maketitle

\setcounter{footnote}{0}  
\begin{abstract}
  Dataset Distillation (DD) seeks to create a condensed dataset that, when used to train a model, enables the model to achieve performance similar to that of a model trained on the entire original dataset. 
  It relieves the model training from processing massive data and thus reduces the computation resources, storage, and time costs. This paper illustrates our solution that ranks 1st in the ECCV-2024 Data Distillation Challenge (track 1).
  Our solution, Modified Difficulty-Aligned Trajectory Matching (M-DATM), introduces two key modifications to the original state-of-the-art method DATM: (1) the soft labels learned by DATM do not achieve one-to-one correspondence with the counterparts generated by the official evaluation script, so we remove the soft labels technique to alleviate such inconsistency; (2) since the removal of soft labels makes it harder for the synthetic dataset to learn late trajectory information, particularly on Tiny ImageNet, we reduce the matching range, allowing the synthetic data to concentrate more on the easier patterns. In the final evaluation, our M-DATM achieved accuracies of 0.4061 and 0.1831 on the CIFAR-100 and Tiny ImageNet datasets, ranking 1st in the Fixed Images Per Class (IPC) Track. \footnote{Codes are available at \href{https://github.com/ChuhaoZhou99/M-DATM}{https://github.com/ChuhaoZhou99/M-DATM}.}

\end{abstract}

\section{Introduction}
\label{sec:intro}

\begin{figure}[tb]
  \centering
  \includegraphics[width=\linewidth]{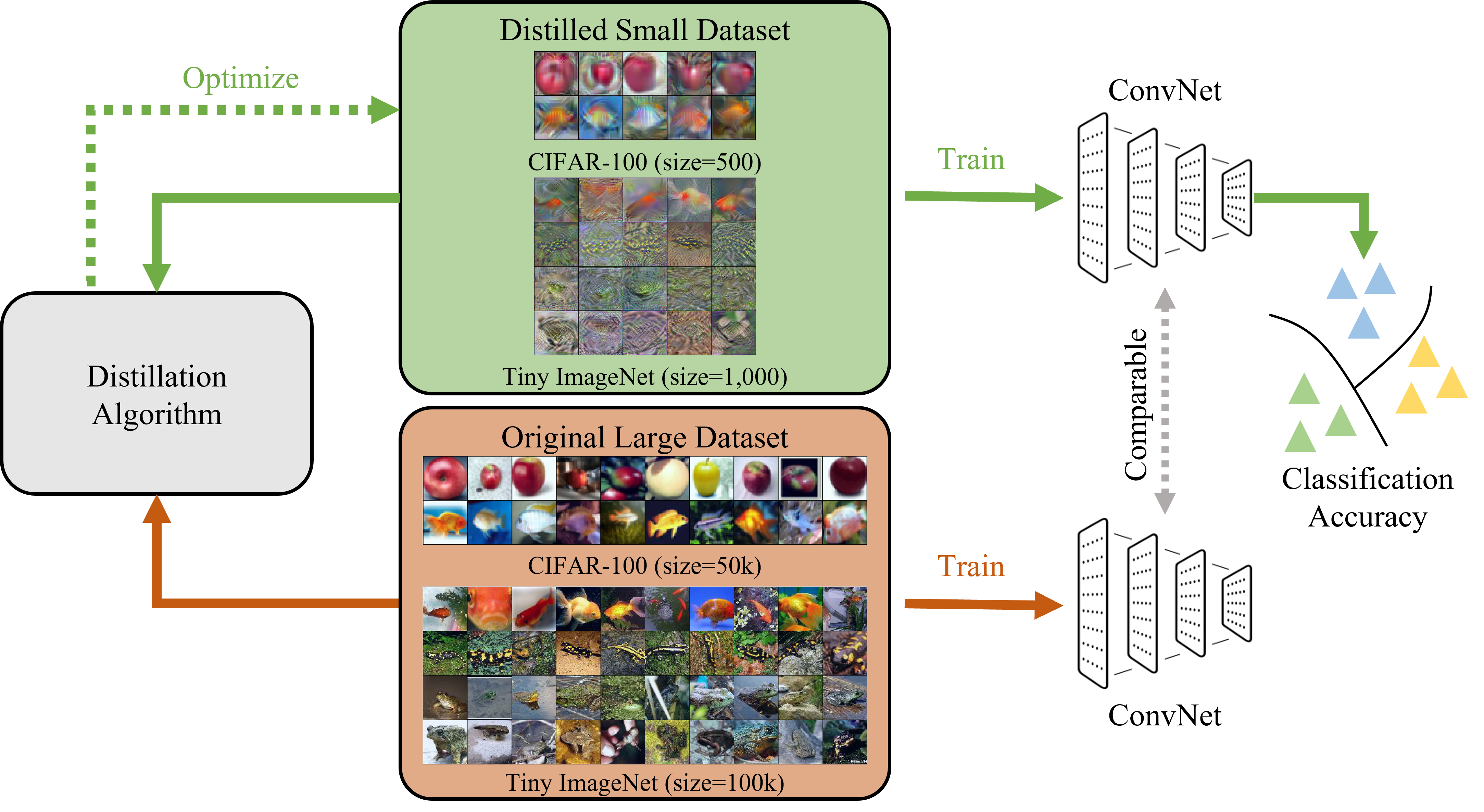}
  \caption{The goal of dataset distillation (DD) challenge. In the challenge, the large original CIFAR-100 (Tiny ImageNet) dataset with size 50K (100K) is distilled to a synthetic small dataset with size 500 (1,000), and the `ConvNet' trained on both datasets are expected to have comparable performances. The classification accuracy serves as the evaluation metric in the challenge.}
  \label{fig:intro}
\end{figure}

Large-scale training data plays a key role in the remarkable success of modern deep learning methods across 
natural language processing~\cite{achiam2023gpt, devlin2018bert, vaswani2017attention}, computer vision~\cite{dosovitskiy2020image, carion2020end} and multi-modal AI~\cite{radford2021learning, li2022blip, liu2023grounding}. However, training models with massive data is extremely resource-intensive in terms of computation, storage, and time, 
which poses a barrier for researchers with limited computational resources. To this end, Dataset Distillation (DD) has been proposed to distill a large-scale dataset into a small synthetic one so that the training effort can be reduced. As shown in~\cref{fig:intro}, the goal of DD is that a model trained on the synthetic small dataset could obtain a comparable performance as a model trained on the original large dataset. The emergence of DD advances data-efficient model training, substantially reducing the costs of the tasks associated with data storage, hyper-parameter tuning, and architectural search~\cite{du2023minimizing}. 

So far, the DD task remains an open problem due to the significant performance gap between results obtained on synthetic datasets and those on real-world counterparts. Many works have been proposed to alleviate the issue from the perspectives of gradient matching~\cite{zhao2021dataset,kim2022dataset, liu2023dream}, distribution matching~\cite{wang2022cafe, zhao2023dataset} and trajectory matching~\cite{cazenavette2022dataset, guotowards,du2023minimizing}. Among them, MTT~\cite{cazenavette2022dataset} serves as the seminal trajectory matching (TM) method that learns synthetic datasets by matching the training trajectory segments (i.e., the time sequences of network parameters) of surrogate models optimized over both the synthetic dataset and the real one. It simultaneously alleviates existing DD methods from (1) being short-sighted (i.e., focusing on single steps) and (2) being difficult to optimize (i.e., modeling the full trajectories), which makes the TM-based methods achieve impressive performance. However, as the size of the synthetic dataset increases, MTT becomes less effective. DATM~\cite{guotowards} further reveals the fact that matching \texttt{early} or \texttt{late} training trajectories will cause the synthetic data to learn \texttt{easy} or \texttt{hard} patterns. Additionally, mismatches between the learning patterns and the capacity of the synthetic dataset (which depends on its size) can degrade the effectiveness of the distillation. Therefore, the difficulty of the learning patterns must be carefully set to align with the capacity of the distilled dataset. 
To address this issue, DATM manages to control the difficulty of learning patterns by only matching the training trajectories within specific ranges. The alignment can then be achieved by selecting the optimal matching ranges that correspond to the size of synthetic datasets. Thanks to the strategy, DATM maintains its effectiveness across both low- and high-IPC (Images Per Class) settings, taking the very first step to lossless dataset distillation.

To promote the development of DD techniques and drive future work, the first DD challenge at ECCV 2024 is held and establishes standard baselines on public datasets, ensuring a fair evaluation of various methods. This report illustrates the ranking 1st solution in this challenge with our revamps and implementation details based on existing state-of-the-art approaches.
In the challenge, we chose the DATM as our baseline model since its flexibility caused by controlling the difficulty of learning patterns allows the DATM to perform well on different datasets. However, two problems have arisen when implementing the DATM to the DD challenge. 
To begin with, the DATM is originally designed to learn soft labels for the synthetic dataset during the distillation. Nevertheless, we find the learned soft labels do not achieve one-to-one correspondence with the labels generated by the official evaluation script that follows a default order. Such \textit{label inconsistency} will make certain synthetic images assigned to incorrect labels during the evaluation, which causes significant performance degradation when the official evaluation script is utilized. Besides, it is observed the DATM consistently obtains poor performance on the Tiny ImageNet dataset. The vanilla DATM attempts to capture relatively hard patterns in Tiny ImageNet by matching the late trajectory information. However, we observe that the learning objective cannot be effectively optimized. Therefore, we hypothesize that it is not reasonable to let DATM focus on the hard patterns in Tiny ImageNet when the size of the synthetic dataset is limited.

In this report, we propose Modified DATM (M-DATM) to tackle the aforementioned issues. To eliminate the \textit{label inconsistency} and ensure the identical labels in the synthetic dataset and those generated during evaluation, M-DATM removes the soft labels technique and directly optimizes the synthetic dataset utilizing labels in default order. However, the removal of soft labels will additionally restrict the information capacity of the synthetic dataset. According to our hypothesis, its ability to capture the late trajectory information (hard patterns) will be further reduced. Therefore, we seek to reduce the difficulty of the learning patterns and let the synthetic dataset concentrate on much easier patterns for effective optimization. Thanks to the property of DATM, it could be easily achieved by adjusting the matching ranges of the training trajectory. In summary, our contributions lie in three folds:

\begin{itemize}
    \item \textbf{Remove the soft labels.} We remove the soft labels technique in M-DATM to ensure consistency between the labels in the synthetic dataset and those generated in the default orders.
    \item \textbf{Adjust the matching range.} We identify the poor performance on Tiny ImageNet results from the difficulty of learning late trajectory information, which is addressed by reducing the matching range.
    \item \textbf{Strong baseline.} Our M-DATM ranks 1st in Track 1 of the ECCV-2024 DD challenge, establishing a strong baseline for future works.
\end{itemize}

\begin{figure}[tb]
  \centering
  \includegraphics[width=\linewidth]{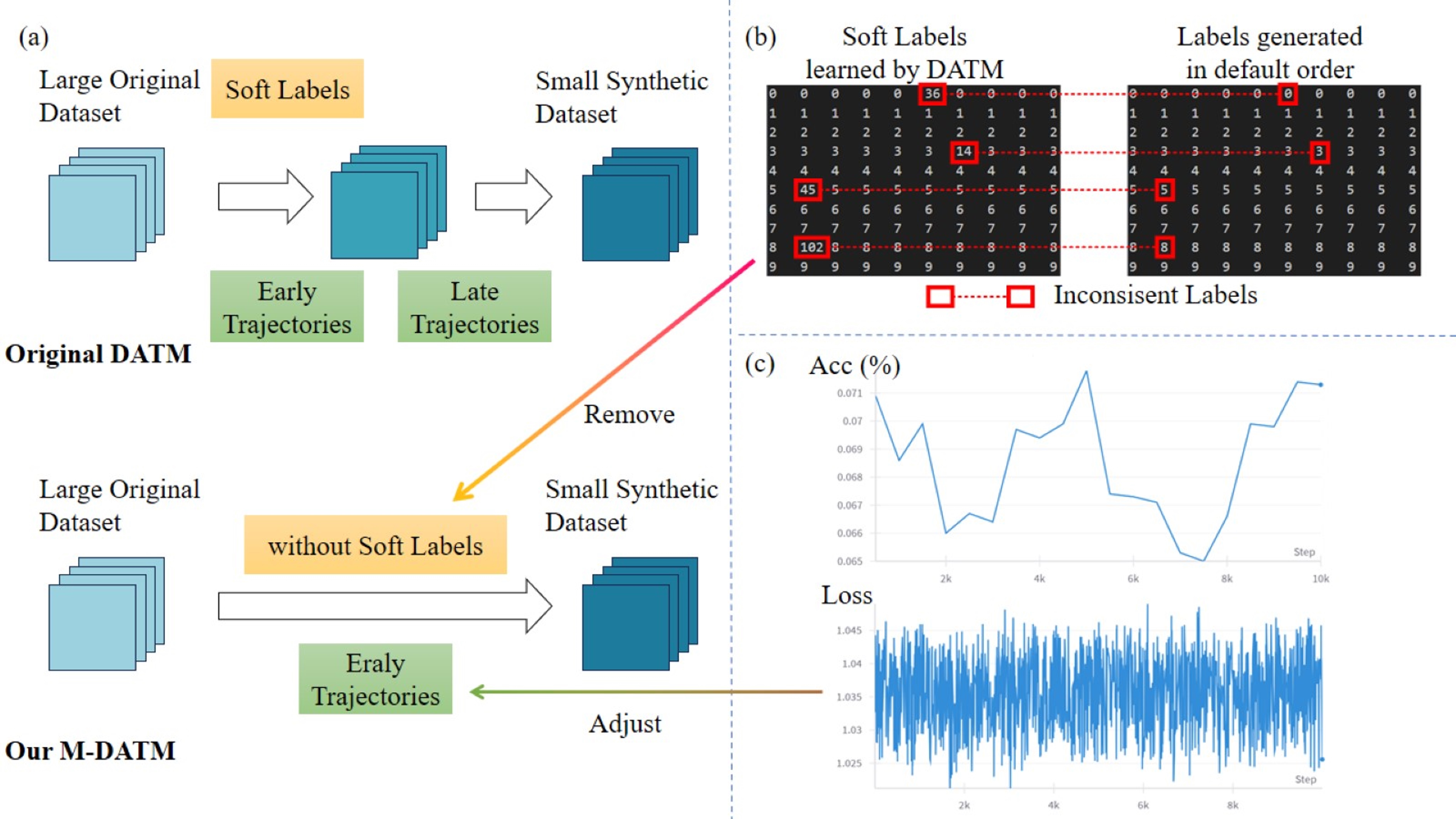}
  \caption{The insights of our M-DATM. (a) Two key modifications in M-DATM: the removal of soft labels technique and the adjustment of the matching ranges. (b) The inconsistency between soft labels learned by DATM and the counterparts generated by the official evaluation script. (c) The DATM could not effectively capture discriminative information through the distillation.}
  \label{fig:method}
\end{figure}

\section{Method}
In this section, we first introduce our baseline model, Difficulty-Aligned Trajectory Matching (DATM). Then the two problems encountered by DATM during the challenge are discussed. Subsequently, we elaborate on two key modifications made to establish the M-DATM, which essentially address label inconsistency and difficulty in capturing late trajectory information.

\subsection{Difficulty-Aligned Trajectory Matching (DATM)}
The goal for dataset distillation is to synthesize a small dataset $\mathcal{D}_{syn}$ that a model trained on $\mathcal{D}_{syn}$ could achieve comparable performance as a model trained on the full, real dataset $\mathcal{D}_{real}$.

For the trajectory matching (TM) methods, the distillation process is conducted by matching the training trajectories of several surrogate models optimized over $\mathcal{D}_{real}$ and $\mathcal{D}_{syn}$. Specifically, the training trajectories obtained from training a surrogate model on $\mathcal{D}_{real}$ are termed as \textit{expert training trajectories}. They are denoted as $\tau^*=\{\theta_t^*\}_0^n$, where the $\theta_t^*$ means the expert network parameters at the training step $t$ and $n$ denotes the total number of training steps for a certain expert. Similarly, the $\hat{\theta}_t$ represents the network parameters trained on the $\mathcal{D}_{syn}$ at the $t$-th training step.

\textbf{Difficulty-Aligned Technique.} During the distillation, the DATM will sample $\theta_t^*$ and $\theta_{t+M}^*$ from the expert training trajectories $\tau^*$ to form the start and target parameters for the matching. The $M$ is a preset hyper-parameter. As mentioned in~\cref{sec:intro}, DATM demonstrates that matching early or late trajectories causes the synthetic data to learn easy or hard patterns. Therefore, the difficulty of the generated patterns could be controlled by restricting the matching range of the trajectories. In detail, a lower bound $T^-$ and an upper bound $T^+$ are separately set to establish a sample range, i.e., only the expert network parameters within $\{\theta_t^*|T^-\leq t \leq T^+\}$ could be sampled for the distillation. As a result, the segment of expert training trajectories utilized for the matching can be formulated as:
\begin{equation}
  \tau^*=\{\textcolor{gray}{\underbrace{\theta_0^*, \theta_1^*,\cdots,}_{\text{too easy}}}\underbrace{\theta_{T^-}^*, \cdots, \theta_{T^+}^*,}_{\text{matching range}}\textcolor{gray}{\underbrace{\cdots,\theta_n^*}_{\text{too hard}}}\}.
  \label{eq:trajectory-segment}
\end{equation}
\textbf{Soft Labels Technique.} The soft labels technique is adopted in DATM to enrich the information capability of the synthetic dataset. To avoid mislabelling, DATM firstly samples a pre-trained model $f_{\theta}^*$ from expert training trajectories. Afterward, all samples in $\mathcal{D}_{real}$ that can be correctly classified by the $f_{\theta}^*$ are selected to form the subset $\mathcal{D}_{sub}$. Then, DATM randomly selected samples from $\mathcal{D}_{sub}$ to initialize the $\mathcal{D}_{syn} = \{(x_i, \hat{y}_i=\text{softmax}(L_i))\}$. Notably, the classification logits $L_i=f_{\theta}^*(x_i)$ are treated as the \textit{soft labels}, which will be updated in each distillation iteration.

Once the matching range is determined, the DATM would optimize the synthetic dataset $\mathcal{D}_{syn}$ by minimizing the distilled loss:
\begin{equation}
  \mathcal{L} = \frac{\|\hat{\theta}_{t+N} - \theta^*_{t+M}\|_2^2}{\| \theta^*_t - \theta^*_{t+M} \|_2^2}
  \label{eq:DATM-distill}
\end{equation}
where $N$ is a preset hyper-parameter that determines the number of training steps on $\mathcal{D}_{syn}$ for each distillation iteration, the $\hat{\theta}_{t+N}$ is obtained in the inner optimization with soft cross-entropy (SCE) loss $l_{soft}$ and the trainable learning rate $\alpha$:
\begin{equation}
  \hat{\theta}_{t+i+1} = \hat{\theta}_{t+i}-\alpha \nabla l_{soft} (\hat{\theta}_{t+i}, \mathcal{D}_{syn}), \text{where} \ \ \hat{\theta}_{t}:=\theta^*_{t}
  \label{eq:DATM-inner-optim}
\end{equation}
where the $\theta^*_t$ is uniformly sampled from the pre-defined matching range and could be formulated as:
\begin{equation}
  \theta^*_t \sim \mathcal{U}(\{ \theta_{T^-}^*, \cdots, \theta_{T}^*\}), \text{where} \ \ T \rightarrow T^+
  \label{eq:DATM-start}
\end{equation}
In~\cref{eq:DATM-start}, $T$ is a floating upper bound that is relatively small at the beginning and will be increased during the distillation process until it reaches $T^+$. In this way, more easy patterns would be sampled at the earlier stage, and the distillation can become more stable.

In each distillation iteration, the $\theta^*_t$ and $\theta^*_{t+M}$ are first sampled from expert training trajectories as start and target parameters. Then, the $\hat{\theta}_{t+N}$ can be obtained by classification with soft labels (~\cref{eq:DATM-inner-optim}). At the end of the distillation iteration, the matching loss can be calculated by~\cref{eq:DATM-distill}. It is then back-propagated to update the synthetic data $x_i$ as well as the soft labels $L_i$, $\mathcal{D}_{syn} = \{(x_i, \hat{y}_i=\text{softmax}(L_i))$\} denotes the targeted synthetic dataset.

\subsection{Modified Difficulty-Aligned Trajectory Matching (M-DATM)}
During the challenge, we identify two problems when applying DATM to the DD task. In this subsection, we first elaborate on these issues, followed by a comprehensive analysis and the corresponding modifications (as shown in \cref{fig:method}(a)) to solve them.

\textbf{Remove Soft Labels.} The first problem we met during the challenge is the performance gap between the evaluation script of DATM and that of the DD challenge. In other words, the test performance obtained from the official evaluation script of the DD challenge is much lower than that obtained from the DATM evaluation script. We have found that the problem is related to the soft labels technique in DATM. As shown by the red squares in~\cref{fig:method}(b), the soft labels learned by DATM do not achieve a precise one-to-one correspondence with the labels generated by the evaluation script (which are generated in the default order). Such \textit{label inconsistency} will cause certain synthetic images to be assigned incorrect labels during evaluation, leading to a performance gap. To this end, we remove the soft labels techniques from DATM and directly optimize the synthetic dataset utilizing labels generated in the default order. Formally, the soft cross-entropy (SCE) loss $l_{soft}$ in~\cref{eq:DATM-inner-optim} is replaced by the standard cross-entropy (CE) loss $l$:
\begin{equation}
  \hat{\theta}_{t+i+1} = \hat{\theta}_{t+i}-\alpha \nabla l (\hat{\theta}_{t+i}, \mathcal{D}_{syn}), \text{where} \ \ \hat{\theta}_{t}:=\theta^*_{t}
  \label{eq:DATM-inner-optim-2}
\end{equation}
where the $\mathcal{D}_{syn} = \{(x_i, y_i=i \mid \text{IPC}$\}, $i$ is the index of each synthetic image, and $\text{IPC}=10$ means the number of images per class. For example, the generated labels for a synthetic dataset with 3 classes and $\text{IPC}=2$ would be $\{ 0,0,1,1,2,2\}$.
With this modification, the synthetic dataset can be directly optimized to meet the requirement of the DD challenge, thereby achieving better performance.

\textbf{Adjust Matching Range.} In the challenge, the other problem for DATM is its poor performance on Tiny ImageNet. Compared to CIFAR-100, Tiny ImageNet dataset contains richer information, with more classes (200 V.S. 100) and a higher resolution ($64 \times 64$ V.S. $32 \times 32$), making it more challenging for existing DD methods. After removing the soft labels technique, we found that the original DATM, which matches a relatively late trajectory on Tiny ImageNet, could not be effectively optimized. As shown in~\cref{fig:method}(c), the loss function and accuracy repeatedly oscillate around the initial point, indicating that the synthetic dataset fails to capture discriminative information during the distillation. With this observation, we conjecture that the removal of soft labels further reduces the information capacity of the synthetic dataset, making it more challenging to capture the late trajectory information (hard patterns). Consequently, we reduce the matching range to $(T^-,T^+)=(0,20)$ to let the synthetic dataset concentrate on easier patterns. This solution has proven effective, achieving the expected performance on Tiny ImageNet. Additionally, we conduct further experiments in~\cref{subsec:exp_reults} to explore the optimal matching ranges for both CIFAR-100 and Tiny ImageNet datasets.

\section{Experiments}
\subsection{Dataset}
The DD challenge utilize two datasets: CIFAR-100~\cite{krizhevsky2009learning} and Tiny ImageNet~\cite{le2015tiny}, both of which are commonly used datasets in dataset distillation literature. The IPC is set to 10 for both datasets in the DD challenge.

\textbf{CIFAR-100.} The CIFAR-100 dataset has 100 classes containing 600 $32 \times 32$ colour images each. There are totally 50,000 images for training and 10,000 images for testing.

\textbf{Tiny ImageNet.} The Tiny ImageNet dataset has 200 classes containing 500 $64\times 64$ images each for training. Each class additionally contains 50 images for validation and 50 images for testing.

\subsection{Implementation Details}
\label{subsec:details}
In this subsection, we provide the detailed settings of the proposed M-DATM, including the distillation process, network, hyper-parameters, and computing resources.

\textbf{Distillation.} We generate expert training trajectories in the same way as FTD~\cite{du2023minimizing}. Consistent with the original DATM~\cite{guotowards}, the distillation process is performed for 10,000 iterations to ensure the convergence. In addition, we remove the ZCA whitening from the DATM and pre-normalized the data utilizing the mean and standard deviation provided by the official evaluation script.

\textbf{Network.} Following the requirements of the DD challenge, we use the default `ConvNet' for expert training trajectories extraction and dataset distillation on both CIFAR-100 and Tiny ImageNet datasets.

\begin{table*}[htb]\small
  \centering
  \caption{Hyper-parameters for different datasets.}
    \resizebox{0.8\textwidth}{!}{
    \begin{tabular}{cc|cccccccc}
    \toprule
    \multirow{2}{*}{Dataset} & \multirow{2}{*}{IPC} & \multirow{2}{*}{$N$} & \multirow{2}{*}{$M$} & \multirow{2}{*}{$T^-$} & \multirow{2}{*}{$T$} & \multirow{2}{*}{$T^+$} & \multirow{2}{*}{Interval} & Synthetic & Learning Rate \\
    &&&&&&&& Batch Size & (Label) \\
    \midrule
    CIFAR-100 & 10 & 40 & 2 & 0 & 15 & 20 & 100 & 1000 & 1000 \\
    \hline
    TI & 10 & 25 & 2 & 0 & 15 & 20 & 250 & 1000 & 10000 \\
    \bottomrule
    \end{tabular}%
  }
  \label{tab:hyper-para}%
\end{table*}%

\textbf{Hyper-parameters.} The hyper-parameters of the M-DATM are reported in~\cref{tab:hyper-para}.

\textbf{Computing resources.} Our experiments are run on 4 NVIDIA A100 GPUs, each with 80 GB memory. The experimental environment is strictly followed the official codes of DATM~\footnote{https://github.com/NUS-HPC-AI-Lab/DATM.}.
\subsection{Main Results}
\label{subsec:exp_reults}
In this subsection, we conduct a comprehensive ablation study on the two modifications of M-DATM. Besides, more experiments are performed to explore the optimal learning patterns for CIFAR-100 and Tiny ImageNet datasets under the DD challenge settings. Notably, all results reported in our experiments are obtained using the official evaluation script. Eventually, visualizations are provided to intuitively show the synthetic datasets learned by matching different patterns.

\begin{table}[H]
  \caption{Ablation Results for M-DATM.}
  \label{tab:ablation}
  \centering

  \begin{tabular}{l|cc}
    \toprule
    Methods & CIFAR-100 & Tiny ImageNet\\
    \midrule
    DATM  & 31.11  & 6.90 \\
    DATM+M1 & 39.90 & 7.10 \\
    DATM+M1+M2 (M-DATM) & 40.61 & 18.31 \\
  \bottomrule
  \end{tabular}
\end{table}

\textbf{Ablation Study.} In this subsection, we conduct ablation studies to show the contribution of the proposed two modifications. Specifically, the `DATM' means our baseline model training on the pre-normalized data as mentioned in~\cref{subsec:details}.  
Removing the soft labels (DATM+M1) leads to significant improvement in performance on CIFAR-100, while performance on Tiny ImageNet remains relatively poor. It's noticeable that the `DATM+M1' only removes the soft labels, but the matching range still follows the default settings of the original DATM. After adjusting the matching range to focus the synthetic dataset on easier patterns (DATM+M1+M2), the performances are improved on both datasets, with a remarkable improvement on Tiny ImageNet (+11.21\%). The ablation study demonstrates that our two modifications are essential and effective. 

\begin{figure}[ht]
  \centering
  \includegraphics[width=0.7\linewidth]{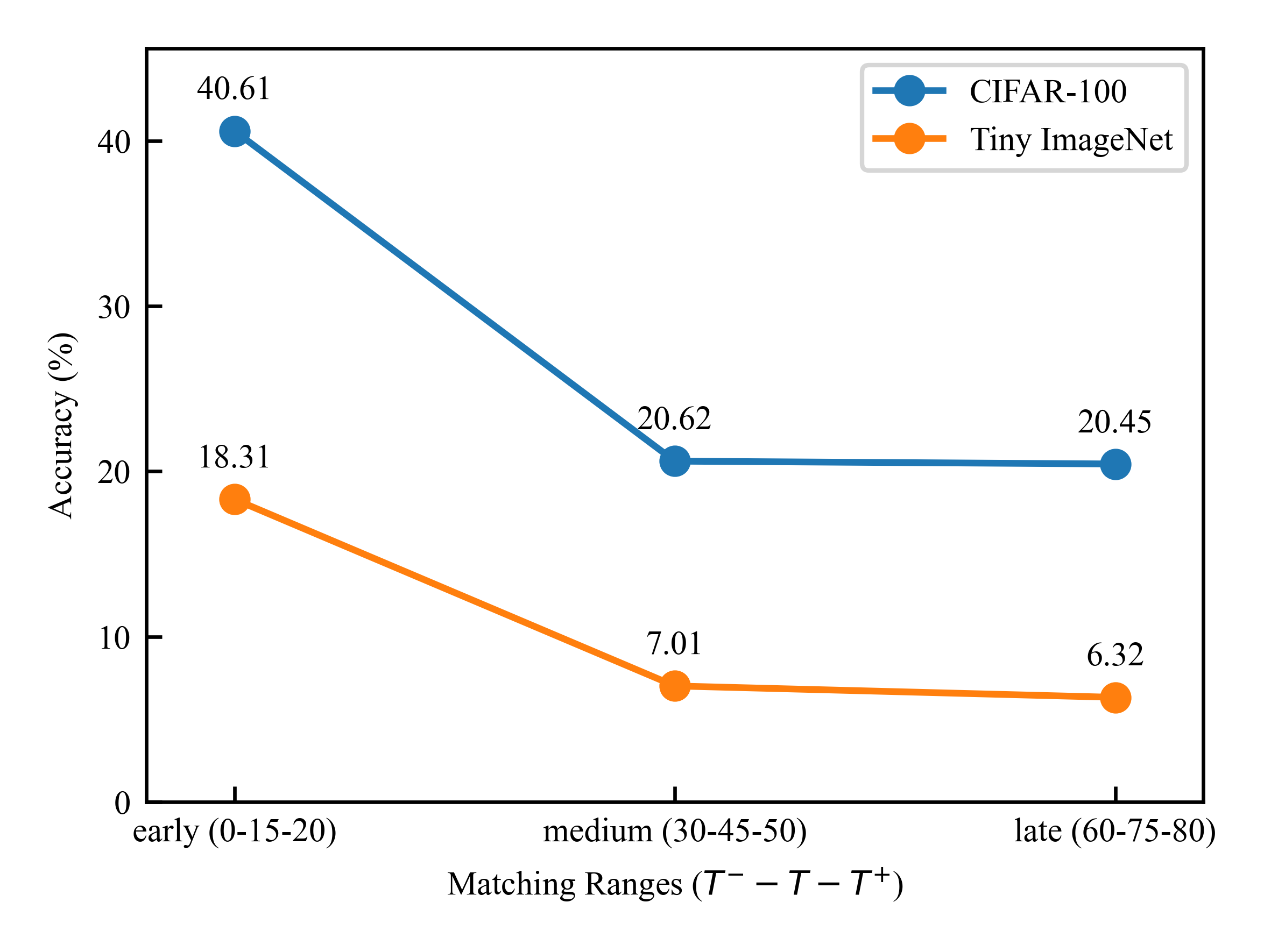}
  \caption{Performances of M-DATM across different matching ranges on CIFAR-100 and Tiny ImageNet.}
  \label{fig:matching-range}
\end{figure}

\textbf{Performances across Different Matching Ranges.} We conduct additional experiments to explore the optimal matching range for both CIFAR-100 and Tiny ImageNet datasets. Specifically, we simply divide the expert training trajectories into three stages: early ($T^-=0$, $T=15$, $T^+=20$), medium ($T^-=30$, $T=45$, $T^+=60$), and late ($T^-=60$, $T=75$, $T^+=80$). The DATM is then evaluated by matching trajectory information at each stage. As shown in~\cref{fig:matching-range}, the best performances for both datasets are achieved at the early stage. The results are consistent with the conclusions drawn from the original DATM. Due to the removal of soft labels and the limited IPC in the DD challenge, the information capacity of the synthetic dataset becomes relatively limited. In this context, making the synthetic dataset focus on the easier pattern, which explains a larger portion of the real data compared to an equal number of hard patterns, could be a better choice.

\begin{figure}[H]
  \centering
  \includegraphics[width=\linewidth]{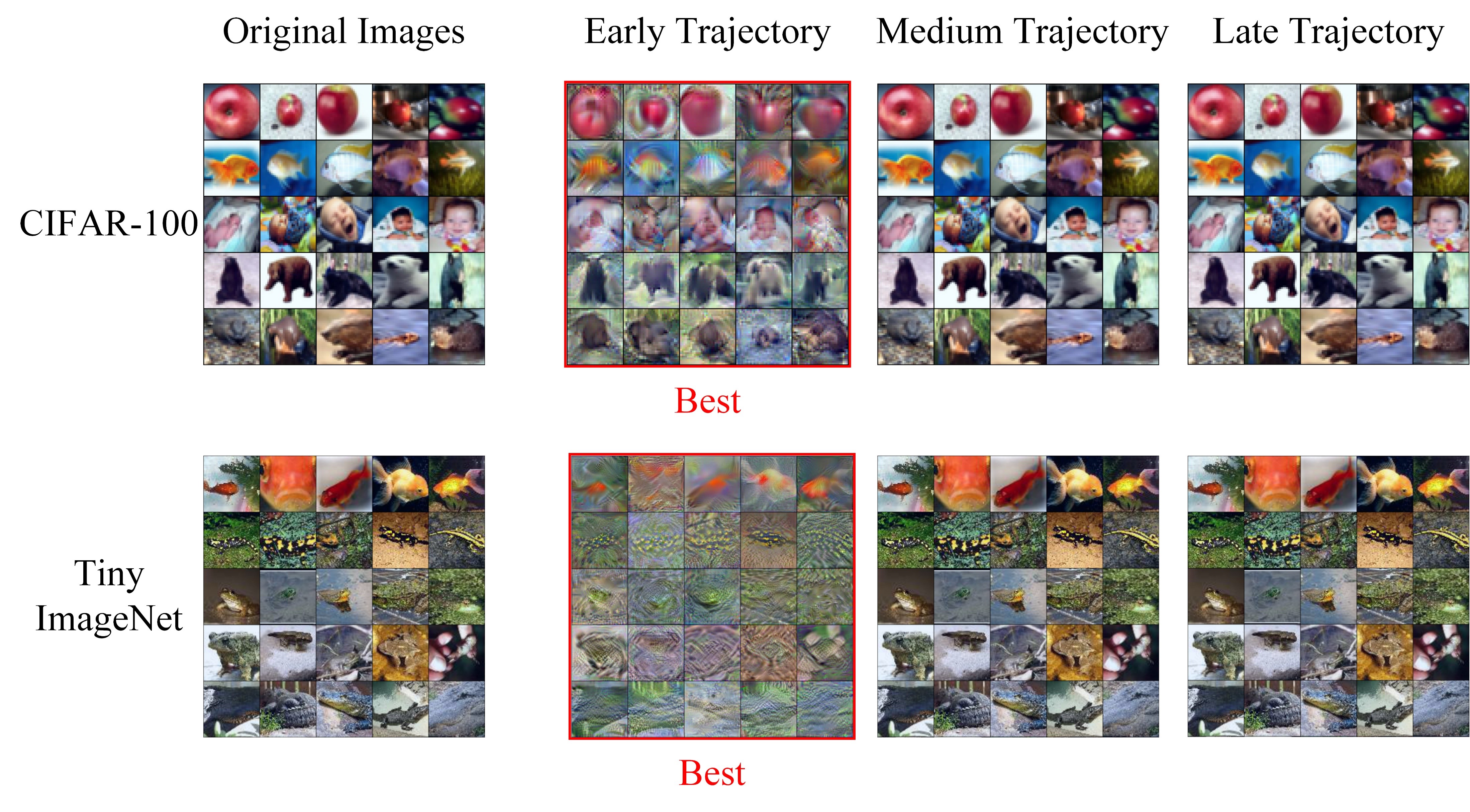}
  \caption{Visualization of the distilled images across different matching ranges on CIFAR-100 and Tiny ImageNet.}
  \label{fig:visualization}
\end{figure}

\subsection{Visualization Analysis} To intuitively show the results of DATM across different matching ranges, we visualize the corresponding synthetic images. As shown in~\cref{fig:visualization}, significant changes are observed in the synthetic images when matching early trajectories, proving that the critical discriminative information is indeed captured through the distillation. In contrast, the limited information capacity of the synthetic dataset prevents it from capturing the medium and late trajectory information, resulting in nearly unchanged synthetic images.

\section{Conclusions}
In this paper, we introduce M-DATM for the DD challenge, incorporating two key modifications to the original DATM method. Specifically, we remove the soft labels technique to ensure one-to-one correspondence between the labels in the synthetic dataset and those generated by the official evaluation script. To improve performance on Tiny ImageNet, we carefully adjusted the matching range to make the synthetic dataset concentrate on the easier patterns. Our M-DATM achieves 1st place in Track 1 of the DD challenge, establishing a strong baseline for future work. Eventually, We thank the organizers for their excellent work in the challenge.


%
%
\bibliographystyle{splncs04}
\bibliography{main}
\end{document}